\documentclass[conference]{IEEEtran}
\IEEEoverridecommandlockouts

\usepackage{times}
\usepackage[numbers]{natbib}
\usepackage{multicol}
\usepackage{multirow}
\usepackage{amsmath}
\usepackage{amsfonts}
\usepackage{booktabs}
\usepackage{siunitx}
\usepackage{graphicx}
\usepackage{amssymb}
\usepackage{pifont}
\usepackage{dsfont}
\usepackage{tabularx}
\usepackage{textcomp}
\usepackage{subcaption}

\usepackage{wrapfig,lipsum,booktabs}
\usepackage{xcolor,colortbl}

\definecolor{citecolor}{HTML}{0071bc}
\usepackage[pagebackref=false,breaklinks=true,letterpaper=true,colorlinks,citecolor=citecolor,bookmarks=false,urlcolor=citecolor]{hyperref}

\definecolor{es-blue}{rgb}{0,0.4,0.8}

\makeatletter
\def\blfootnote{\xdef\@thefnmark{}\@footnotetext}
\makeatother

\begin{document}

\title{Expressive Whole-Body Control for \\ Humanoid Robots}

\author{\authorblockN{Xuxin Cheng\authorrefmark{1}, Yandong Ji\authorrefmark{1}, Junming Chen, Ruihan Yang, Ge Yang, Xiaolong Wang
}
\vspace{0.05in}
\authorblockA{UC San Diego}
{\color{es-blue}{\texttt{\url{https://expressive-humanoid.github.io}}}}
}

\twocolumn[{%
\renewcommand\twocolumn[1][]{#1}%
\maketitle
\begin{center}
    \centering
    \captionsetup{type=figure}
    \includegraphics[width=1.0\textwidth]{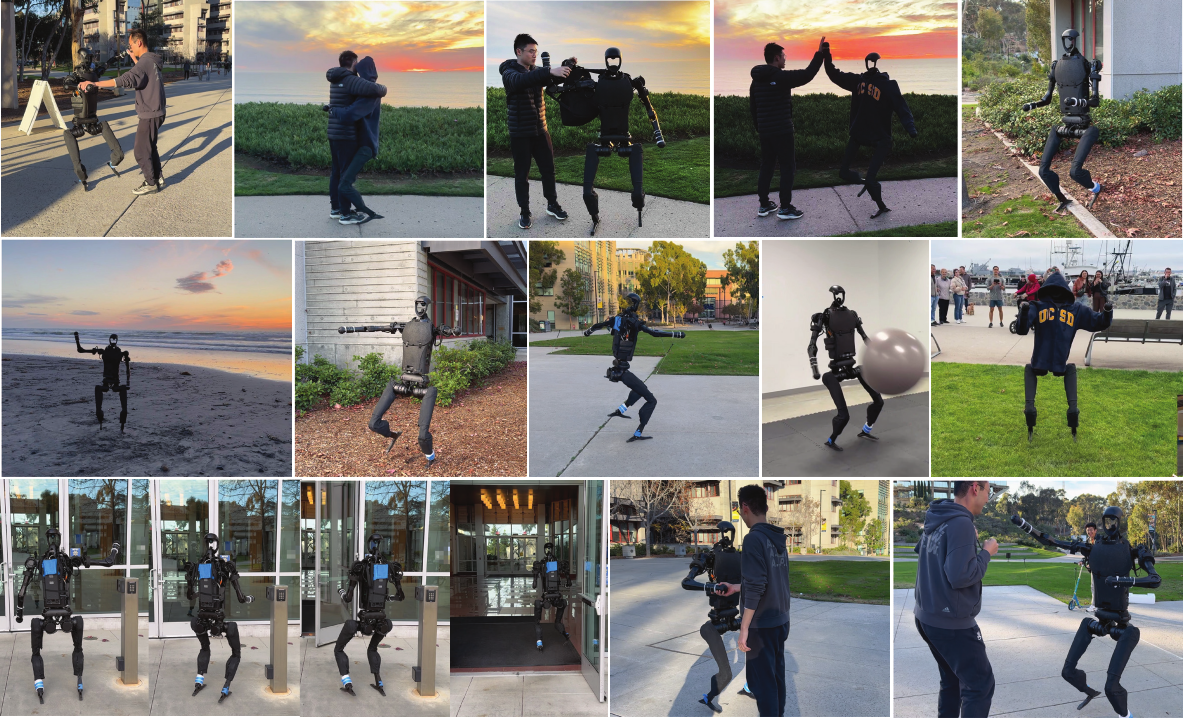}
    \caption{Our Robot demonstrates diverse and expressive whole-body movements in different scenarios. Top Row: The robot is dancing, hugging and slapping hands with a human. Middle Row: The robot is able to walk on different terrains including gravel and wood chip paths, inclined concrete paths, grass, and curbsides with various expressions like zombie walk, exaggerated stride or waving. Bottom Left: The robot is able to use a waving gesture to open a wave-sensing door. Bottom Right: The robot is shaking hands and provoking. }
    \vspace{-2pt}
    \label{fig:teaser}
\end{center}
}]
{\blfootnote{{$^{*}$ The first two authors contributed equally.}}}

\begin{abstract}
Can we enable humanoid robots to generate rich, diverse, and expressive motions in the real world? We propose to learn a whole-body control policy on a human-sized robot to mimic human motions as realistic as possible. To train such a policy, we leverage the large-scale human motion capture data from the graphics community in a Reinforcement Learning framework. However, directly performing imitation learning with the motion capture dataset would not work on the real humanoid robot, given the large gap in degrees of freedom and physical capabilities. Our method \textbf{Ex}pressive Whole-\textbf{Body} Control (\textbf{ExBody}) tackles this problem by encouraging the upper humanoid body to imitate a reference motion, while relaxing the imitation constraint on its two legs and only requiring them to follow a given velocity robustly. With training in simulation and Sim2Real transfer, our policy can control a humanoid robot to walk in different styles, shake hands with humans, and even dance with a human in the real world. We conduct extensive studies and comparisons on diverse motions in both simulation and the real world to show the effectiveness of our approach.

\end{abstract}

\IEEEpeerreviewmaketitle

\section{Introduction}
\begin{figure*}[t!]
    \centering
    \includegraphics[width=\linewidth]{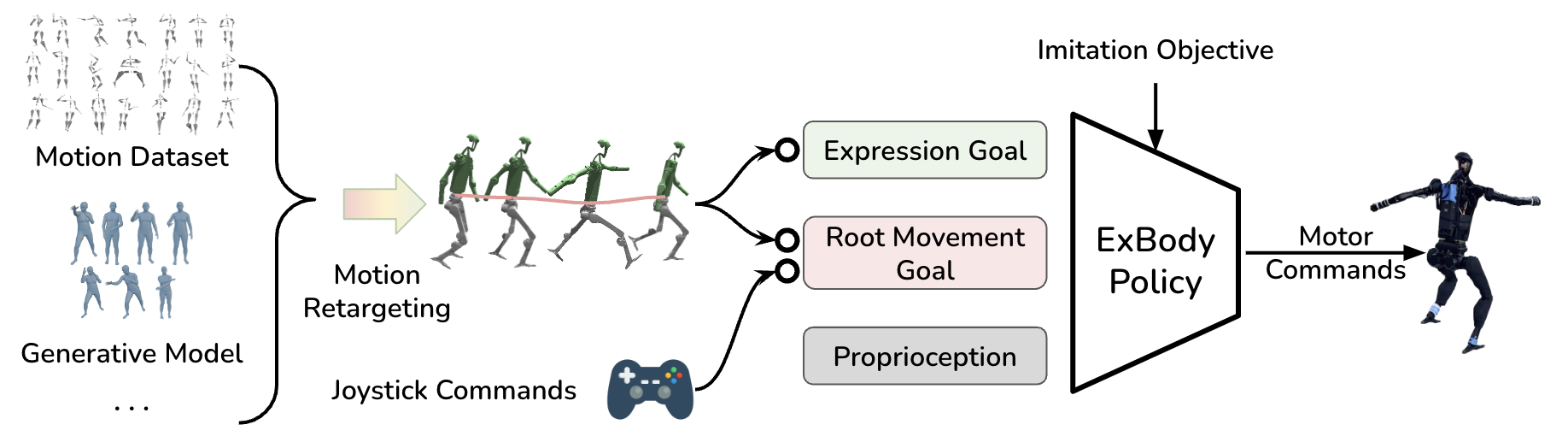}
    \caption{Overview of our framework. Our framework is able to train on data from various sources such as static human motion datasets, generative models, video to pose models that are widely available. After motion retargeting, we acquire a repertoire of motion clips that are compatible with our robot's kinematic structure. We extract expression goal $\mathcal{\mathbf{g}}^e$ and root movement goal $\mathcal{\mathbf{g}}^m$ from the rich features from retargeted motion clips as the goal of our goal-conditioned RL objective. The root movement goal $\mathcal{\mathbf{g}}^m$ can also be intuitively given by joystick commands, enabling convenient deployment in the real world.}
    \label{fig:method}
\end{figure*}

When we think of robots, we often begin by considering what kinds of tasks they can accomplish for us. Roboticists typically work under this framework, and formulate control as optimizing for a specific cost function or task objective. When applied to robots that resemble our house pets such as quadruped robot dogs, or humans, whole-body control methods on both of these two form factors tend to produce singular motion patterns that lack grace and personality  --- oftentimes a by-product of the additional constraints we have to add to make optimization or learning easier. In contrast, motions from actual humans and animals are rich, diverse, and expressive of their intent or emotional valence. In other words, there exists a large subspace of motion control that is not described by common objectives such as body velocity, heading, and gait patterns. What would it take to build robots that can generate, and perform diverse whole-body motions that are as expressive as humans?

In this paper, we tackle the problem of learning a whole-body motion control policy for a human-sized robot that can match human motions in its expressivity and richness. We do so by combining large-scale human motion capture data from the graphics community with deep Reinforcement Learning (RL) in a simulated environment, to produce a whole-body controller that can be deployed directly on the real robot. We illustrate the expressiveness of our controller in Fig.~\ref{fig:teaser}, and show that the robot is sufficiently compliant and robust that it can hold hands and dance with a person. 

Our work benefits from prior research from the computer graphics community on physics-based character animation~\cite{llobera2023physics}, and from the robotics community on using deep reinforcement learning to produce robust locomotion policy on various legged robots~\cite{kumar2021rma,agarwal2023legged}. In our study, we found that although physics-based character animation produces natural-looking reactive control policies that look good in a virtual setting, such results often involve large actuator gains in the range of \(60 \text{kg/m}\) that are one magnitude larger than what is feasible with current hardware. We also found that human reference motion often involves a lot more degrees of freedom (DoF) than the robot hardware. For example, the physics-based animation can use much more DoF (e.g., \(69 \text{DoF}\)~\cite{Luo2023PerpetualHC}) compared to a real-world robot (e.g., \(19 \text{DoF}\) on a Unitree H1 robot). These two factors make the direct transfer of graphics techniques onto the real robot infeasible.

\begin{figure*}[t!]
    \centering
    \begin{subfigure}[t]{0.24\linewidth}
    \centering
    \includegraphics[width=\linewidth]{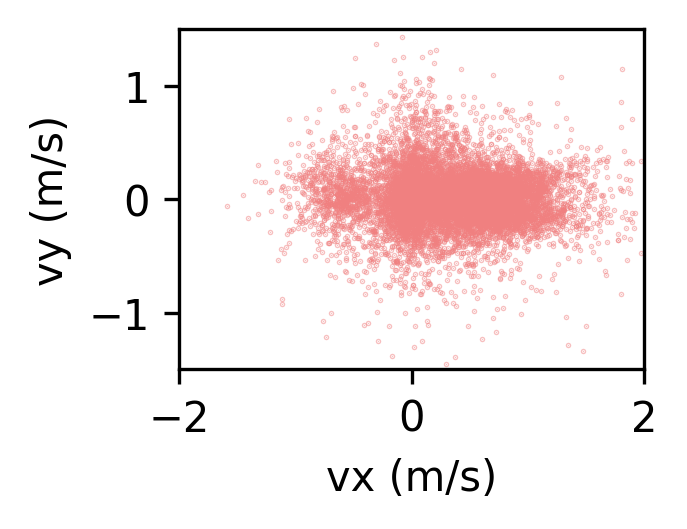}
    \caption{}
    \label{fig:vis1}
    \end{subfigure}
    \begin{subfigure}[t]{0.25\linewidth}
    \centering
    \includegraphics[width=\linewidth]{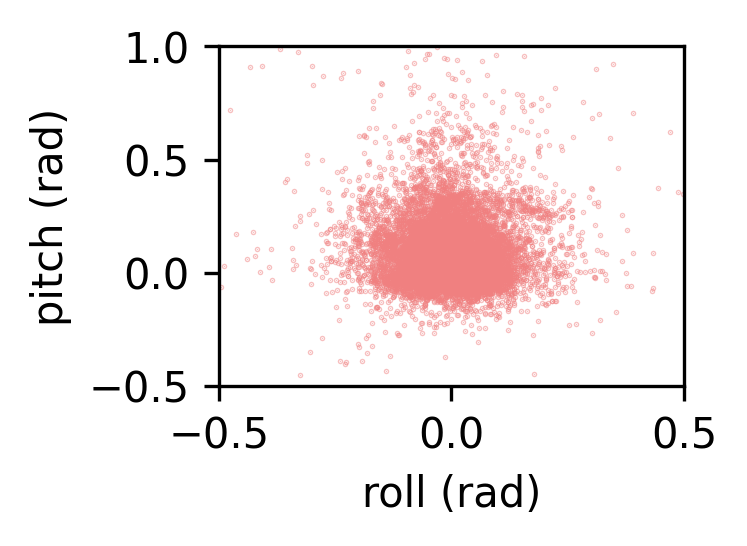}
    \caption{}
    \label{fig:vis2}
    \end{subfigure}
    \begin{subfigure}[t]{0.24\linewidth}
    \centering
    \includegraphics[width=\linewidth]{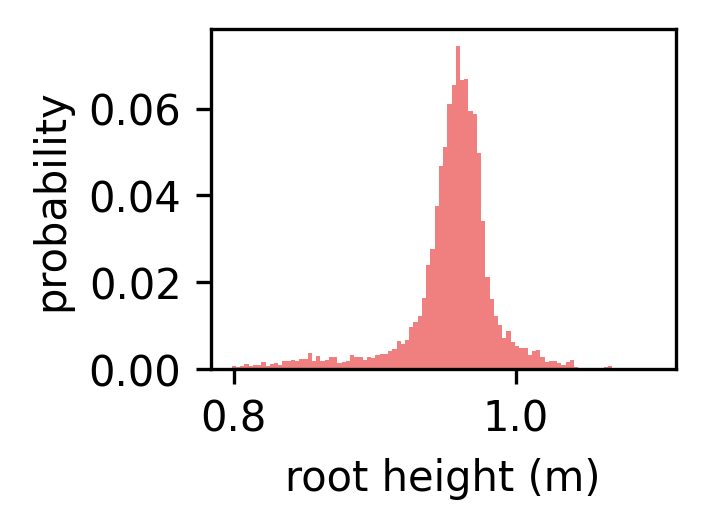}
    \caption{}
    \label{fig:vis3}
    \end{subfigure}
    \begin{subfigure}[t]{0.24\linewidth}
    \centering
    \includegraphics[width=\linewidth]{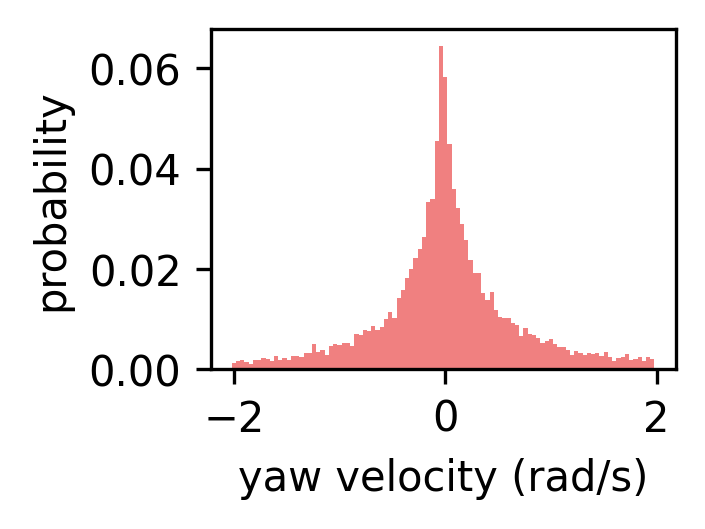}
    \caption{}
    \label{fig:vis4}
    \end{subfigure}
    
    \caption{Dataset visualization of our training data from CMU MoCap. We sample all the motion clips at an incremental of 1s. The resulting number of plotting data points are 1338. We can observe the bias of the distribution from human motions. Such distributions are proven to help policy learning in Sec. \ref{sec:results}.}
    \label{fig:data_vis}
\end{figure*}

Our key idea is to NOT mimic exactly the same as the reference motion. We propose to train a novel controller that takes both a reference motion and a root movement command as inputs for real humanoid robot control. We call our approach \textbf{Ex}pressive Whole-\textbf{Body} Control (\textbf{ExBody}). During training with RL, we encourage the upper body of the humanoid robot to imitate diverse human motions for expressiveness, while relaxing the motion imitation term for its two legs. Concretely, the reward function for the legged locomotion is designed for following the root movement commands robustly provided by the reference motion instead of matching each exact joint angle. We train our policy in highly randomized challenging terrains in simulation. This not only allows robust sim2real transfer but also learns a policy that does not just ``repeat'' the given motion. The user can command the humanoid robot to move at different speeds, turning in different directions on diverse terrains, and reproduce the reference motion on the upper body at the same time. As shown in Fig.~\ref{fig:teaser}, we can command our robot to dance with a human, waving and shaking hands while walking, or walking like a mummy on diverse terrains.

We adopt the Unitree H1 robot in both simulation and real-world experiments. To learn from diverse human motions, we utilize the CMU MoCap dataset (around 780 reference motions). Such richness not only enables more expressive humanoid motion but also more robust walking. Our evaluation shows the upper body motions and diverse moving velocity augment the training data and provide efficient guidance in training. We also compare our method with applying more imitation constraints on legged motion in both simulation and the real world and show our approach that relaxes the constraints indeed leads to better and more robust results. To the best of our knowledge, our work is the first work on learning-based real-world humanoid control with diverse motions. While our current results focus on expressive humanoid control, we hope our approach can also shed some light on studying generalizable humanoid whole-body manipulation and navigation. 

\begin{table}[h]
    \centering
    \begin{tabular}{c|c|c|c}
    \toprule
     Metrics & Mimic WBC \newline (Ours) & PHC \cite{Luo2023PerpetualHC} & ASE \cite{2022-TOG-ASE} \\
    \midrule
    DoFs & 19 & 69 & 37 \\
    Number of Motion Clips & 780 & 11000 & 187 \\
    Total Time of Motions (h) & 3.7 & 40 & 0.5 \\
    Real Robot &\checkmark &$\times$ & $\times$ \\
    Single Network &\checkmark &$\times$ & \checkmark\\
    Linear Velocities Obs & $\times$ & \checkmark & \checkmark  \\
    Keypoint Positions Obs& $\times$ &  \checkmark &\checkmark \\
    Robot Height Obs & $\times$ & $\times$ & \checkmark
    \end{tabular}
    \caption{Comparisons with physics-based character animation works. In PHC, the policy observes the Linear velocities and keypoint positions of each rigid body, while in ASE linear velocities are for the root only. PHC and ASE both observe privileged states that are not available on the real robot.}
    \vspace{-10pt}
    \label{tab:comparisons}
\end{table}

\section{Problem Formulation}

We consider humanoid motion control as learning a goal-conditioned motor policy \(\pi: \mathcal G \times \mathcal S \mapsto \mathcal A\), where \(\mathcal G\) is the goal space that specifies the behavior, \(\mathcal S\) is the observation space, and \(\mathcal A\) is the action space that contains the joint positions and torque. We assume in the rest of this paper, without loss of generality, that the observation and action space are given by the H1 humanoid robot design. However, our proposed approach should generalize to similar body forms that differ in the exact number of actuated degrees of freedom.

\paragraph{Command-conditioned Locomotion Control} We aim to produce a robust control policy for the Unitree H1 hardware that can be commanded by the linear velocity \(\mathbf v \in \mathbb R^3\), body pose in terms of row/pitch/yaw \( rpy \in \mathbb R^3\) and the body height  \(h\) measured at the root link. 
Formally, the goal space for root movement control \(\mathcal G^m = \langle \mathbf{v}, rpy, h \rangle\). The observation \(\mathcal S\) includes the robot's current proprioception information $s_t=[ \omega_{t}, r_t, p_t, \Delta y, q_t, \dot{q}_t, \mathbf{a}_{t-1}]^T$. $\omega_t$ is the robot root's angular velocity, $r_t, p_t$ is roll and pitch. Note that the policy does not observe the current velocity $\mathbf{v}$, and the absolute body height $h$ and the current yaw angle $y_t$ because these are privileged information for the real robot (see Tab.~\ref{tab:comparisons}). 
We let the policy observe the difference between current and desired yaw angle $\Delta y = y_t - y$ to convert the global quantity to a local frame that can be intuitively commanded at deployment time. 
The actions $\mathbf{a}_t\in\mathbb{R}^{19}$ is the target position of joint-level proportional-derivative (PD) controllers. The PD controllers compute the torque for each motor with the specified PD gains \(k^i_p\) and damping coefficient \(k^i_d\).

\paragraph{\textbf{Ex}pressive Whole-\textbf{Body} Control} We extend the command-conditioned locomotion control to include descriptions of the robot's movement that are not captured by root pose and velocity in \(\mathcal G^m\). We formulate this as the more general goal space \(\mathcal G = \mathcal{G}^e \times \mathcal G^m\), where the expression target \(\mathbf{g}^e \sim \mathcal G^e\) includes the desired joint angles and various 3D keypoint locations of the body.

Specifically, in this work, we work with a relaxed problem where we exclude the joints and key points from the lower half of the body from \(\mathcal G^e\). This is because the robot has a different body plan from humans, and including these low-body features from human motion capture data tends to over-constrain the problem and lead to brittle, and poorly performing control policies. Formally, for the rest of this paper, we work with  \(\mathcal G^e = \langle \mathbf q, \mathbf p \rangle\), where \(\mathbf q \in \mathbb R^9\) are the joint positions of the nine actuators of the upper body, and \(\mathbf p \in \mathbb R^{18}\) are the 3D key points of the two shoulders, two elbows, and the left and right hands. The goal of expressive whole-body control (ExBody) is to simultaneously track both the root movement goal (for the whole body) \(\mathbf g^m\sim \mathcal G^m\), as well as the target expression goal (for upper body) \(\mathbf g^e\sim \mathcal G^e\).

\begin{figure*}[t]
\centering
\includegraphics[width=0.9\linewidth]{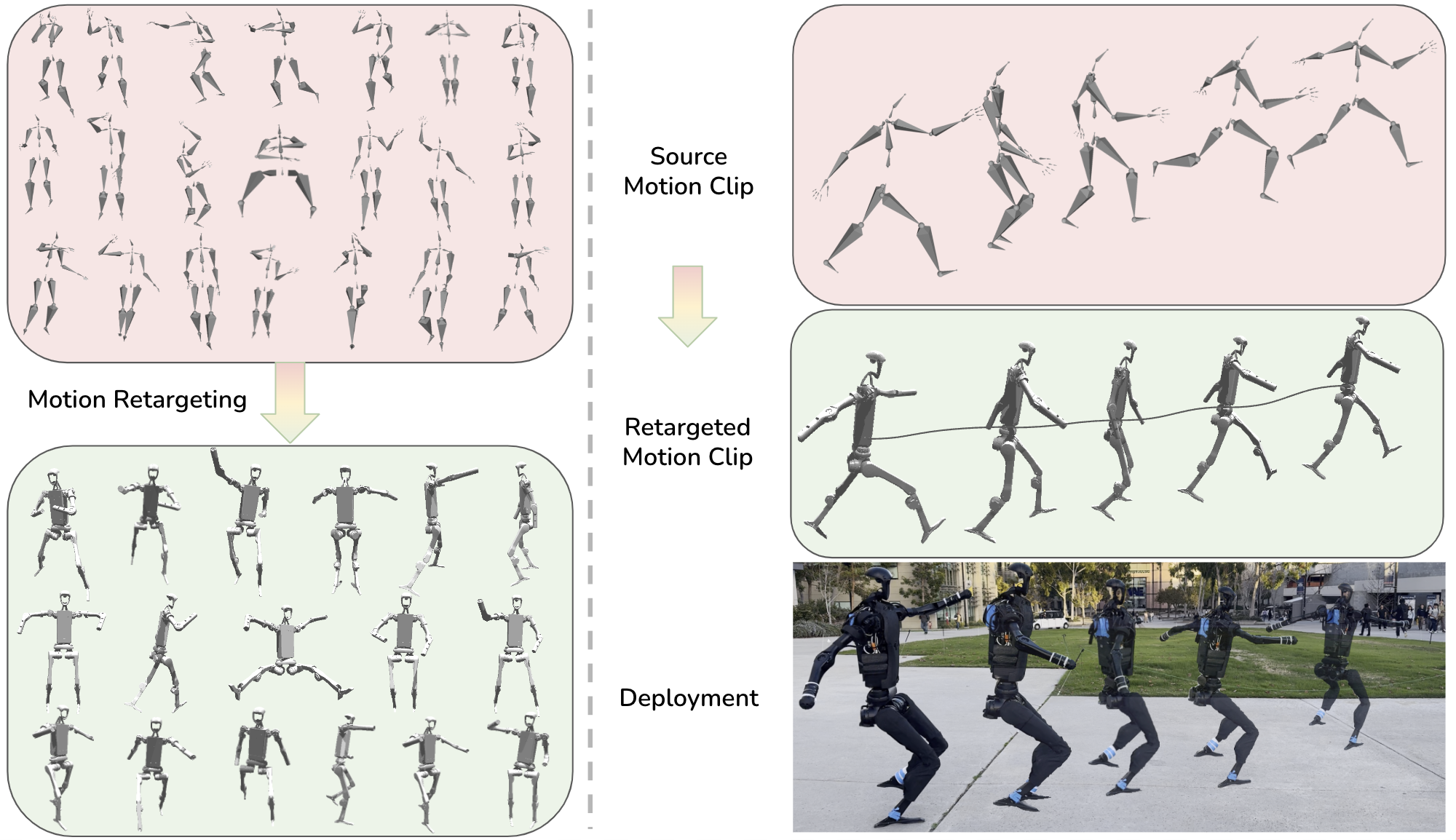}
\caption{Left: During training, we extract a large repertoire of retargeted motion clips and train our ExBody policy. Right: During deployment, we can replay motion that can come from a variety of sources such as static motion datasets, diffusion models, or video-to-skeleton models. For Unitree H1, the robot we use, the shoulder and hip joints have three perpendicular DoFs. Other joints are 1 DoF each. There are 19 DoFs in total. We also notice that some of the retargeted motions exhibit exaggerated movement with robot's lower body, which is why we use ExBody to make it transferrable.}
\vspace{-0.2in}
\label{fig:retarget}
\end{figure*}
\section{Expressive Whole-Body Control}

We present Expressive Whole-Body Control (ExBody), our approach for achieving expressive and robust motion control on a humanoid robot as shown in Fig. \ref{fig:method}. In the following sections, we cover the key components of this approach, including strategies for curating and retargeting human motion capture data to humanoid robot hardware, and using some of these prior knowledge to improve the RL training procedure.

\subsection{Strategies for Curating Human Behavior Data}
In our research, we selectively used a portion of the CMU MoCap dataset, excluding motions involving physical interactions with others, heavy objects, or rough terrain. This was done semi-automatically, as our framework cannot realistically implement motions with significant environmental interactions. The resulting motions are in Tab.~\ref{tab:dataset}.

\begin{table}[h]
    \centering
    \begin{tabular}{c|c|c|c}
    \toprule
        & Category & Clips & Length $(s)$ \\
    \midrule
        \multirow{6}{*}{Training}
        & Walk & 546 & 9076.6 \\
        & Dance & 78 & 1552.3\\
        & Basketball & 36 & 766.1 \\
        & Punch & 20 & 800.0 \\
        & Others & 100 & 1188.0 \\
        & Total & 780  & 13383.0 \\
    \midrule
        \multirow{12}{*}{Real-World Test} & Punch & 1 & 18.9  \\
        & Wave Hello & 1  & 5.0\\
        & Mummy Walk & 1 & 22.5 \\
        & Zombie Walk & 1 & 13.0\\
        & Walk, Exaggerated Stride  & 1 & 2.5 \\
        & High Five & 1  & 3.3\\
        & Basketball Signals & 1 & 32.6  \\
        & Adjust Hair & 1 & 9.6 \\
        & Drinking from Bottle & 1 & 15.2 \\
        & Direct Traffic & 1 & 39.3 \\
        & Hand Signal & 1  & 32.2\\
        & Russian Dance & 1 & 8.2  \\
        & Total & 11 & 202.3  \\
        
    \midrule
    \multirow{3}{*}{\shortstack{Additional \\ Realworld Test \\(Diffusion \cite{tevet2022human})}}
        & Boxing & 1 & 4.0 \\
        & Hug & 1 & 4.0 \\
        & Shake Hands & 1 & 4.0 \\
    \end{tabular}
    \caption{The details of our dataset. We select a subset from CMU MoCap dataset for training, and test on various expressive motions in sim and the real world. }
    \label{tab:dataset}
\end{table}

We plot the distribution of root movement goal $\mathbf{g}^e$ in Fig. \ref{fig:data_vis}. The yaw angle in the world frame does not have too much meaning because during training all our observations are in the robot's local frame. So we visualize the yaw angular velocity instead.
From Fig. \ref{fig:vis1}, we can observe that the motions we choose cover walking forward, backward, and sideways, and are biased towards walking forward and symmetrical on walking sideways. From Fig. \ref{fig:vis2}, we observe a chestnut-shaped distribution with minimal movement along the roll angle, because humans do not usually incline the body sideways. However, in terms of pitch, it is biased toward bending forward instead of bending backward. And as the pitch angle gets larger the roll angle gets smaller in distribution, which resonates with human bias. From Fig. \ref{fig:vis3}, we observe that the robot's height is narrowly centered around its nominal height and has small variations. From Fig. \ref{fig:vis4} we can see that our turning motion is pretty balanced and there are turning left and right motions. Unlike \cite{fu2022learning}, which randomly samples points using a spherical coordinate frame, and then checks if the points are under the ground or have a collision with the robot itself, our approach from large human data naturally has samples that generally do not violate such constraints. Even if there are some collisions with the robot itself after retargeting, the RL will avoid it via collision penalization. We show in Sec. \ref{sec:results} that this prior distribution actually helps with policy learning a lot compared with manually designed sample space.

\subsection{Motion Retargeting to Hardware}
In consideration of the distinct morphological differences between the H1 robot and humans, we adapt the human motion data to the robot's framework by straightforwardly mapping local joint rotations onto the robot's skeleton. We use the Unitree H1 robot\cite{unitree2024website} as our platform with a total mass of around $51.5kg$ and a height of around $1.8m$. The Unitree H1 robot has 19 DoFs. 
The shoulder and hip joints have 3 revolute joint motors connected perpendicularly, thus equivalent to a spherical joint usually used in human motion datasets \cite{AMASS:2019, Guo_2022_CVPR}. During retargeting, we consider the 3 hip or shoulder joints as 1 spherical joint. After retargeting, we remap the spherical joint which is represented by a normalized quaternion \( \mathbf{q^i_m} = (q_x, q_y, q_z, q_w) \) to the original joint angle of 3 revolute joints $\mathbf{m}=[m_1, m_2, m_3]\in \mathbb{R}^3$ by exponential mapping. To achieve this we first convert the quaternion $\mathbf{q}$ to the form of axis angle:
\begin{align*}
        \theta = 2 \arccos(q_w), ~
        \mathbf{a} = \frac{1}{\sqrt{1 - q_w^2}} \begin{pmatrix} q_x \\ q_y \\ q_z \end{pmatrix}
    \end{align*}
    where $\mathbf{a}$ is the rotation axis and $\theta$ is the rotation angle. For small angles, the last axis is used if \( \sqrt{1 - q_w^2} \) is close to zero. Then the mapped angle is just simply $\mathbf{m} = \theta \mathbf{a}$, where $\mathbf{m}=[q^i, q^j, q^k]$ is 3 corresponding DoFs in $\mathbf{q}$.
For 1D joints, namely the elbow, torso, knee, and ankle, we take the rotation angle projected onto the corresponding rotation axis of the 1D joints. In Fig. \ref{fig:retarget}, we show the diverse motions both for the original dataset and the retargeted ones. We can see that although the retargeted data loses some DoFs with hardware constraints, it is still able to keep the important expressions from the original data.

\subsection{Guiding State Initialization from Human Mocap Data}
We use massively parallel simulation to train our RL policy with Isaac Gym \cite{makoviychuk2021isaac, rudin2022learning}. We randomly sample an initial state $\mathbf{g}=[\mathbf{g}^e, \mathbf{g}^m]$ from the motion dataset for each environment in simulation and set its state to the sampled state during initialization or resetting. We show by extensive experiments how this random initialization helps with policy learning.

With the diverse goal state $\mathbf{g}$ distribution as shown in Fig.~\ref{fig:data_vis} and the corresponding tracking rewards, ExBody is able to produce diverse root movement and diverse arm expressions while still maintaining the balance via the lower body without mimicking rewards. Our policy can make the robot walk forward/backward, sideways, turn yaw, vary root height, adjust roll, pitch, etc.

\subsection{Rewards}
In each step, the reward from the environment consists of expression goal, root movement goal tracking, and regularization terms derived from \cite{rudin2022learning}.
Imitation rewards are detailed in Tab. ~\ref{tab:rewards_detailed}, where $\mathbf{q}_\text{ref} \in \mathbb R^9$ is reference position of the upper body joints, $\mathbf{p}_\text{ref} \in \mathbb R^{18}$ is reference position of the upper body keypoints, $\mathbf{v}_\text{ref}$ is reference body velocity, $\boldsymbol{\Omega}_\text{ref}^{\phi\theta}$ and $\boldsymbol{\Omega}^{\phi\theta}$ are reference and actual body roll and pitch. Refer to supplementary for regularization rewards.

\begin{table}[h]
\centering
\normalsize
\begin{tabular}{@{}llr@{}}
\toprule
Term & Expression & Weight \\ \midrule
\multicolumn{3}{c}{Expression Goal $G^e$} \\ \midrule
DoF Position & exp$(-0.7 |\mathbf{q}_{\text{ref}}-\mathbf{q}|$) & 3.0 \\
                                  Keypoint Position & exp$(-|\mathbf{p}_{\text{ref}}-\mathbf{p}|$ & 2.0 \\ \midrule
\multicolumn{3}{c}{Root Movement Goal $G^m$} \\ \midrule
                                  Linear Velocity & exp$(-4.0 |\mathbf{v}_\text{ref}-\mathbf{v}|)$ & 6.0 \\
                                  Roll \& Pitch & exp$(-|\boldsymbol{\Omega}_\text{ref}^{\phi\theta}-\boldsymbol{\Omega}^{\phi\theta}|)$ & 1.0 \\
                                  Yaw & exp$(-|\Delta y|)$ & 1.0 \\
\midrule
\end{tabular}
\caption{Expressive Rewards Specification}
\vspace{-0.2in}
\label{tab:rewards_detailed}
\end{table}

\section{Results}
\label{sec:results}
\begin{figure*}[t]
\centering
    \begin{subfigure}[t]{\linewidth}
    \centering
    \includegraphics[width=\linewidth]{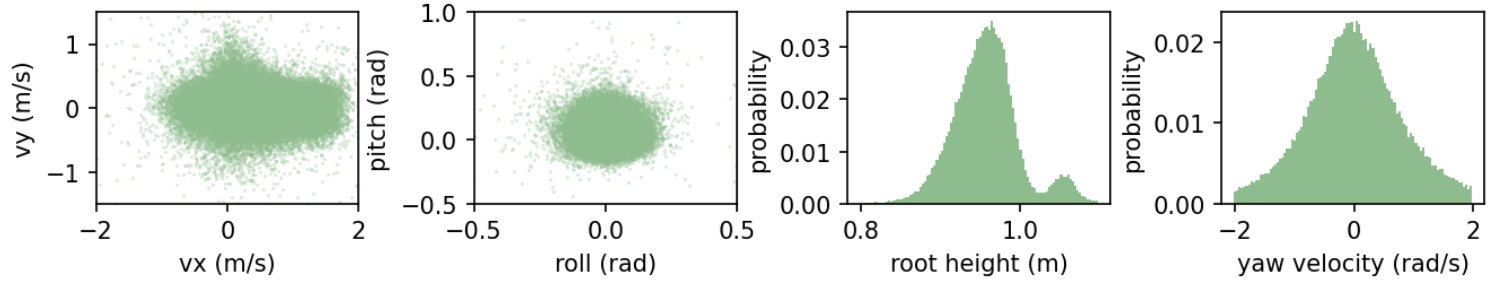}
    \caption{Dataset sampling $\mathbf{g}^m$}
    \label{fig:motion_sample}
\end{subfigure}
\begin{subfigure}[t]{\linewidth}
    \centering
    \includegraphics[width=\linewidth]{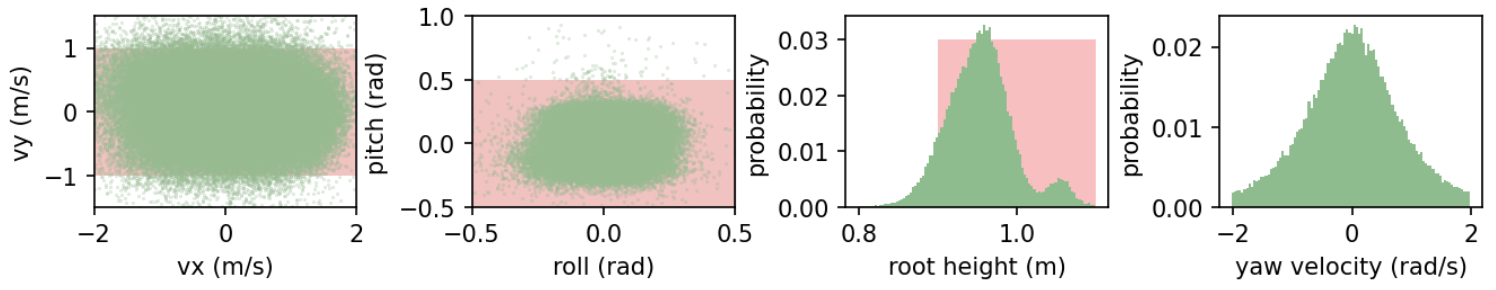}
    \caption{Random sampling $\mathbf{g}^m$}
    \label{fig:random_sample}
\end{subfigure}
\caption{Policy's state distribution under different sampling strategies. The green dots are the policy rollout's states. For dataset sampling, we record 20 data points for 4096 environments with randomly sampled arm trajectories from our training set. For random sampling, he red shade represents the randomly sampled $\mathbf{g}^m$ range. For yaw velocity, we do not sample the command, because the policy observes the difference between the desired and actual yaw, and does not explicitly track the angular velocity. The second peak in root height is the initialization bias. }
\label{fig:ours_rand}
\vspace{-0.2in}
\end{figure*}

In this section we aim to answer the following questions through extensive experiments both in sim and the real world: 
\begin{itemize}
    \item How well does ExBody perform on tracking $\mathbf{g}^e$ and $\mathbf{g}^m$?
    \item How does learning from large datasets help policy exploration and robustness? 
    \item Why do the state of the art approaches in computer graphics for physics based character control not work well in the real robot case and why do we need ExBody? 
\end{itemize}

Our baselines are as follows:
\begin{itemize}
    \item \textbf{ExBody + AMP}: This baseline uses an AMP reward to encourage the policy's transitions to be similar to those in the retargeted dataset.
    \item \textbf{ExBody +AMP NoReg}: We remove the regularization terms in our reward formulations and see if AMP reward itself is able to handle the regularization of the imitation learning problem with such a large dataset.
    \item \textbf{Random Sample}: Randomly sample root movement goals $\mathbf{g}^m$ with the range shown in Fig. \ref{fig:random_sample}.
    \item \textbf{No RSI}: Initialize the environment with default DoF positions and root states instead of sampling from the motion dataset.
    \item \textbf{Full body tracking}: Instead of tracking only the upper body with $\mathcal{G}^e$, the objective is to track the joint angles and 3D key points for the entire body including hips, knees, and ankles.
    
\end{itemize}
Our metrics are as follows:
\begin{itemize}
    \item \textbf{Mean Episode Linear Velocity Tracking Reward} (MELV)
    \item \textbf{Mean episode roll pitch tracking reward} (MERP)
    \item \textbf{Mean episode lengths} (MEL)
    \item \textbf{Mean episode key body tracking reward} (MEK)
    
\end{itemize}

\noindent\textbf{How well does ExBody perform on tracking $\mathbf{g}^m$?}
In Fig. \ref{fig:motion_sample}, we show that our trained ExBody can produce similar root movement state distribution as shown in the training dataset in Fig. \ref{fig:data_vis}. The other way to put it, the policy successfully learned to map a few intuitive root movement commands to appropriate motor commands, and executing these motor commands in a physical world will result in tracking the root movement commands. However, our ExBody is trained on motion clips, which have temporal correlations in terms of the goal state $\mathbf{g}$. That is to say, the robot is commanded to replay a certain trajectory that has meaning. What about the randomly sampled commands? If the policy can handle these commands, we are able to intuitively control the robot in the real world with a joystick, without worrying about finding the correct motion clips and replay them. We show that our trained ExBody policy can generalize to any input trajectories other than those in the dataset. To do this, we randomly sample the root state commands and record the policy's rollout distribution. We can see from Fig. \ref{fig:ours_rand}, that the policy covers most areas of the sampled velocity commands, which is larger than the training set in Fig. \ref{fig:data_vis}. For roll and pitch commands, the policy's state distribution is smaller than the sampled commands. We speculate this is due to the robot's self-property (upper body too heavy, etc). 
The reason why we do not give a number for tracking error is that without comparison, it has no meaning, so qualitatively seeing the data distribution is more straightforward and intuitive.

\begin{figure}[ht]
    \centering
    \includegraphics[width=0.7\linewidth]{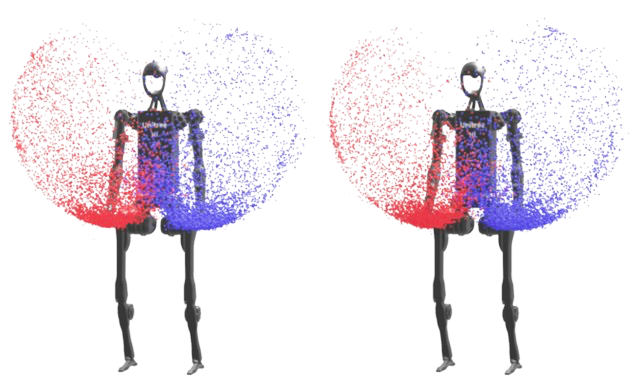}
    \caption{We sample 10,000 points of hand positions relative to the robot. Left: retargeted motion dataset. Right: learned ExBody policy rollouts. The upper body movement from the dataset forms a natural distribution for learning. }
    \label{fig:hand_dist}
    \vspace{-0.2in}
\end{figure}

\noindent\textbf{How well does ExBody perform on tracking $\mathbf{g}^e$ ?}
Similarly, we render the samples of end-effectors (hands) positions relative to the robot to show a nearly identical distribution of reference motion and learned policy as shown in Fig. \ref{fig:hand_dist}.

\begin{figure}[h]
    \centering
    \includegraphics[width=1\linewidth]{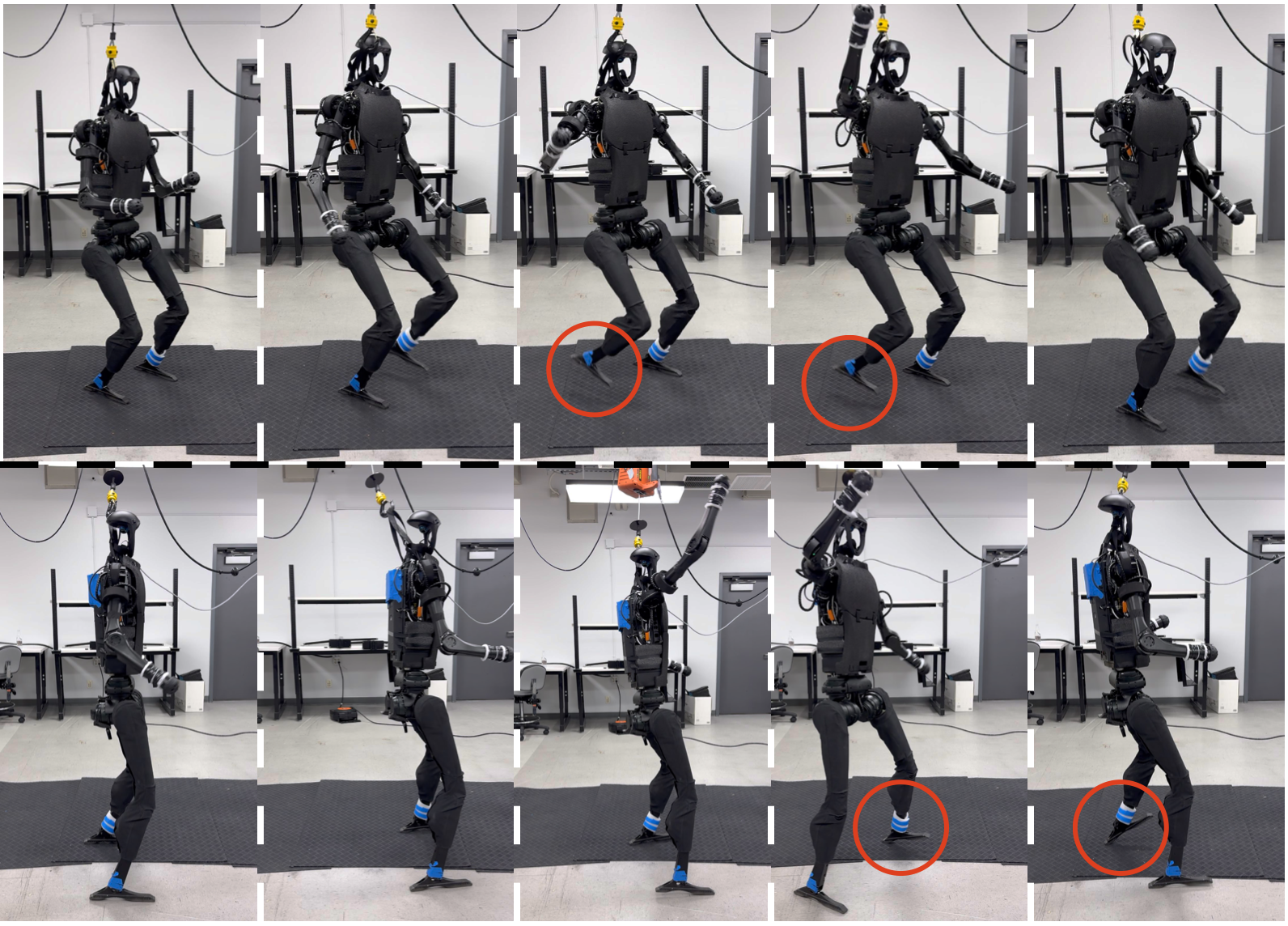}
    \caption{H1 robot doing a High Five in the real world. Top Row: ExBody only (Ours) walks with more bent knees and has more foot height clearance. Bottom Row: ExBody + AMP tends to walk in a straight-leg way and has less foot height clearance during walking. }
    \label{fig:high5}
    \vspace{-0.2in}
\end{figure}

\begin{figure}[ht]
    \centering
    \includegraphics[width=\linewidth]{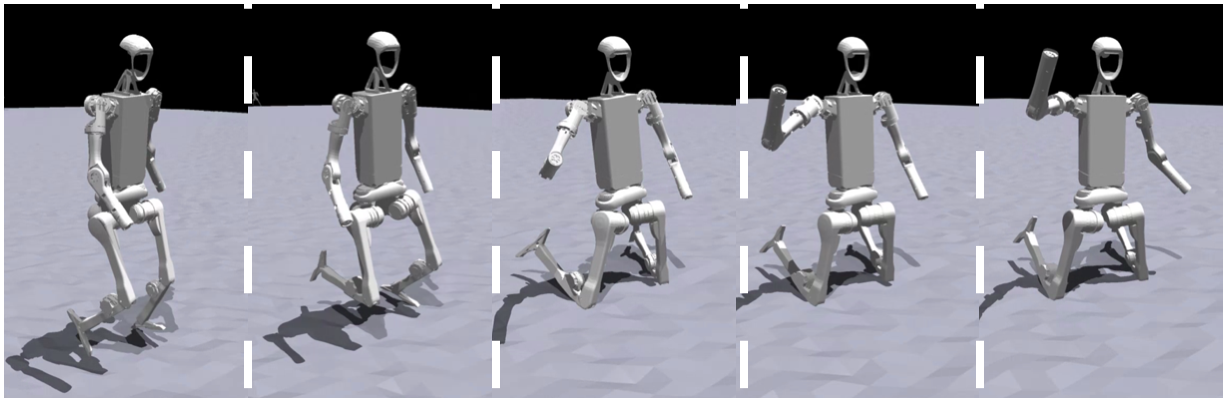}
    \caption{Random Sampling $\mathcal{\mathbf{g}}^m$ results in a behavior that the policy immediately kneels down after initialization, trying to be as stable as possible while ignoring the root movement goal $\mathbf{g}^m$.}
    \label{fig:comp_rand}
    \vspace{-0.2in}
\end{figure}

\noindent\textbf{Why should we learn from large data?}
\begin{table*}[t!]
    \centering
    \normalsize
    \begin{tabularx}{\textwidth}{p{1.6in}XXXXXXXX}
        \toprule 
        \multirow{2}{*}{Baselines} & \multicolumn{4}{c}{Motion Sample} & \multicolumn{4}{c}{Random Sample} \\
        \cmidrule(lr){2-5} 
        \cmidrule(lr){6-9}
         & MEL$\uparrow$  & MELV$\uparrow$ & MERP$\uparrow$ & MEK$\uparrow$ & MEL  & MELV & MERP & MEK \\
        \midrule
         ExBody (Ours)  & 16.87 & \textbf{318.67} & 754.92 & \textbf{659.78} &
                13.51 & \textbf{132.14} & 523.79 & 483.67\\
        ExBody + AMP & \textbf{17.28} & 205.60 & \textbf{765.85} & 635.51 
         & 15.59 & 95.11 & 583.82 & \textbf{544.59}\\
         ExBody + AMP NoReg &16.16 & 87.83 & 714.74 & 561.56 
         & 15.40 & 36.76 & 584.23 & 515.53\\
         No RSI & 0.23 & 0.63 & 10.09 & 7.25 
         & 0.22 & 0.10 & 7.41 & 7.15\\
         Random Sample & 16.50 & 181.85 & 704.73 & 326.66 &
         16.37 & 38.51 & \textbf{586.83} & 324.10 \\
         Full Body Tracking & 13.28 & 246.11 & 584.40 & 397.25
         & 10.76 & 76.46 & 407.88 & 284.69\\
         
         \bottomrule
    \end{tabularx}
\caption{Comparisons with baselines. We sample 10,000 trajectories with 4096 environments in simulation and report their mean episode metrics. Motion Sample means we sample $\mathcal{\mathbf{g}}^m$ from retargeted motions. Random Sample means we uniformly sample $\mathcal{\mathbf{g}}^m$ in Fig. \ref{fig:data_vis}.}
\vspace{-0.2in}
    \label{tab:sim_results}
\end{table*}
We compare with baselines to show that our approach ExBody is superior compared with other design choices. Traditionally, RL-based robust locomotion for legged robots is trained either through reward engineering or from a limited set of reference motions. In our work, we show the advantage of learning robust Whole-Body control for humanoid robots from large motion datasets. As shown in Tab. \ref{tab:sim_results}, our method achieves the best linear velocity tracking performance (MELV). The benefit largely comes from RSI, where we initialize the robot to different states that encourage exploration. No RSI is not able to discover proper positive reward states before learning to suicide as soon as possible to avoid negative rewards in simulation. The Random Sample baseline's behavior is a kneel-down motion for all the goals as shown in Fig. \ref{fig:comp_rand}, taking advantage of the environment. It gives up $\mathbf{g}^m$ completely and focuses on $\mathbf{g}^e$. It has a similar MEL score with the motion sample and a higher MEL score with the Random Sample (the training distribution), meaning that kneeling on the ground is more robust. We speculate that the motion dataset offers a more advantageous distribution of $\mathcal{G}^m$, which in turn facilitates the policy learning process. For example, many motions started from standing in place and gradually started to walking, creating a natural curriculum for the policy to learn.

\noindent\textbf{Why does not ExBody  do full DoF tracking?}
Due to the limited torque, DoFs of the real robot, we design ExBody to only mimic the arm motions $\mathbf{g}^e\sim \mathcal{G}^e$ while the whole-body's objective is to track root movement goals $\mathbf{g}^m\sim \mathcal{G}^m$. We show in Tab. \ref{tab:sim_results} that tracking the Whole-Body expressions will result in reduced performance with all metrics. The robot's lower body movements exhibit numerous artifacts, notably that while the reference motion is designed for a single step, the robot executes multiple steps in an attempt to stabilize.

\noindent\textbf{Comparisons with adversarial methods. }We also compare with the adversarial methods that can serve as a regularizer on top of our method \cite{Luo2023PerpetualHC, luo2023universal, escontrela2022adversarial}. Our method plus an AMP regularizer demonstrates better results in terms of $MEL$ and $MERP$. But just like the Random Sample baseline, it is at a great cost of linear velocity tracking ($MELV$). 
The policy generated by AMP often results in a gait with less knee flexion and inadequate foot clearance, leading to toes that tilt non-horizontally and a tendency to stumble while walking, as illustrated in Fig. \ref{fig:high5}. These characteristics could pose challenges for sim-to-real transfer.
We also tested a series of motions in the real world as shown in Tab. \ref{tab:real_result} and recorded their roll and pitch variations as an indicator of how stable the policy is. We can see that Ours + AMP has more shaking than ours. Our method has a stable stepping gait while the AMP one tries to use straight legs and stand in place, which results in significant stumbling and feet artifacts shown in Fig. \ref{fig:high5}.
ExBody + AMP NoReg tries to replace the regularization terms in Tab. \ref{tab:rewards_detailed}. However, it has even worse performance, demonstrating a high-frequency jittery movement that is not feasible for sim-to-real transfer, indicating for such a complex system, AMP reward itself is not sufficient. 

\begin{figure}[t]
    \centering
    \includegraphics[width=1\linewidth]{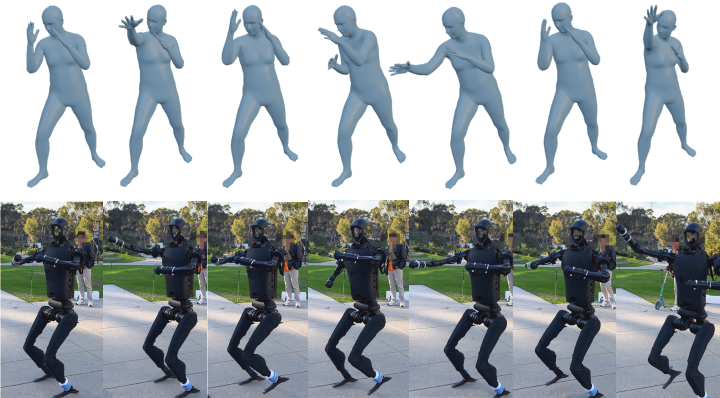}
    \caption{Text2Motion trajectories replay. A motion sequence is prompted offline with the input "a man mimics boxing punches" through MDM \cite{tevet2022human}. Our robot presents robust, responsive, and precise tracking performance. }
    \label{fig:punch}
    \vspace{-0.2in}
\end{figure}

\noindent\textbf{Results in Real World}
We extensively evaluate our framework in the real world, with Fig.~\ref{fig:teaser} and Fig.~\ref{fig:high5} showcasing the replaying of CMU MoCap trajectories, and Fig.~\ref{fig:punch} evaluating text prompting diffusion pipeline. Notably, as walking gaits are not predefined during the training phase, our robot autonomously learns to synchronize its stepping frequency with the upper body poses during motion playback. For additional  demonstrations, please refer to the supplementary materials, where more result videos are available.

\section{Related Work}
\noindent\textbf{Whole-Body Control with Legged Robots}
Legged robots often need to coordinate the entire body to complete some tasks or do some motions such as dancing, reaching for a far object, etc, which were previously primarily achieved by dynamics modeling and control \citep{miura1984dynamic,yin2007simbicon, hutter2016anymal,moro2019whole,dariush2008whole, kajita20013d, westervelt2003hybrid}. However, for a humanoid robot that has a high degree of freedom \citep{grizzle2009mabel, kato1973development, hirai1998development, chignoli2021humanoid, BostonDynamics2024, AgilityRobotics2024}, it will require substantial engineering and modeling \citep{sreenath2011compliant} and are sensitive to real-world dynamics changes. Recent learning-based methods \cite{Fu2023-ve,ito2022efficient, cheng2023legmanip, jeon2023learning, schwarke2023curiosity, ji2023dribblebot, ji2022hierarchical} achieved whole-body locomotion and manipulation for a quadruped robot. These advances also enable better learning-based humanoid control \cite{kumar2023words, tang2023humanmimic, li2023robust, seo2023deep}. However, most of the studies focus more on the locomotion side or learning a relatively small dataset. Different from all previous works, our work enables whole-body control for expressive motions on a human-sized robot in the real world. 

\noindent\textbf{Legged Locomotion}
Blind-legged locomotion across challenging terrains has been widely studied, via reward specification \cite{margolis2022rapid, kumar2021rma, fu2021minimizing, fu2022learning}, via imitation learning \cite{Escontrela22arXiv_AMP_in_real} and gait heuristics \cite{li2021reinforcement, siekmann2021blind}. Vision-based locomotion has achieved great successes traversing stairs \cite{agarwal2023legged, yang2023neural, margolis2021learning, duan2023learning}, conquering parkour obstacles \cite{zhuang2023robot, cheng2023parkour}, manipulating boxes \cite{dao2023sim}. However, these works have not fully taken advantage of demonstration data. Even works utilizing re-targeted animal motions or pre-optimized trajectories still leverage a very small dataset \cite{Peng2020-ty, wang2023amp, Escontrela22arXiv_AMP_in_real, fuchioka2023opt, yang2023generalized}, while our framework can benefit from learning with a large-scale motion dataset.

\noindent\textbf{Physics-based Character Animation}
Whole-body humanoid motion control has been widely studied in the realm of computer graphics, where the goal is to generate realistic character behaviors. Adversarial methods such as \cite{peng2021amp, 2022-TOG-ASE, tessler2023calm, InterPhysHassan2023} suffer from mode collapse as the motions get more and more. Peng et al. \cite{2022-TOG-ASE} used a unit sphere latent space to represent the 187 motions. However, it stills suffers from mode collapse and utilized additional skill discovery objective. Imitation-based methods \cite{ScaDiver, wang2020unicon, zhang2023vid2player3d, 2018-TOG-deepMimic} alleviate this problem by decoupling control and motion generation, where a general motion tracking controller is trained to track any motions and a motion generator outputs motions to track. These works demonstrated successful transfer to real robot quadrupeds \cite{RoboImitationPeng20, Escontrela22arXiv_AMP_in_real}.  \cite{ScaDiver} separate the entire CMU MoCap data into several clusters and train mixture-of-expert policies to reproduce physically plausible controllers for the entire dataset. Luo et al. \cite{Luo2023PerpetualHC} used a similar idea by progressively assigning new networks to learn new motions. However, these methods are hard to transfer to the real humanoid robot because of the unrealistic character model (SMPL humanoid \cite{SMPL:2015} has a total of 69 DoFs with 23 actuated spherical joints and each joint has 3 DoFs, there is usually no torque limit ), privileged information used in the simulation (world coordinates of robots, velocities, etc) demonstrated in Tab. \ref{tab:comparisons}. Thus we propose {Ex}pressive Whole-{Body} Control to address this problem. ExBody relaxes the lower body tracking objective and uses a whole-body root movement goal instead. While considering the capability of our robot, we select a subset of motions that includes mainly walking and everyday behaviors and expressions and we use only one single network for all the motions.

\begin{table}[t]
    \centering
    \begin{tabular}{c|c|c}
    \toprule
        Motions & Ours & Ours+AMP \\ \midrule
Walk, Exaggerated Stride &0.054 &  0.087 \\
Zombie Walk &0.072 &0.11\\
Wave Hello &0.062 &0.095 \\
Walk Happily &0.037 &0.074 \\
Punch &0.052 &0.055 \\
Direct Traffic, Wave, Point &0.037 &0.094 \\
Highfive &0.04 &0.084 \\
Basketball Signals &0.045 &0.081 \\
Adjust Hair Walk &0.042 &0.09 \\
Russian Dance &0.063 &0.1 \\
Mummy Walk &0.064 &0.086 \\ \midrule
Boxing &0.075 &0.068 \\
Hug &0.037 &0.086 \\
Shake Hand &0.036 &0.099  \\
\midrule
        Mean & 0.051 & 0.087  \\
    \end{tabular}
    \caption{We report the mean absolute roll and pitch angle for a 10-second test in the real world for each motion.}
    \label{tab:real_result}
    \vspace{-0.2in}
\end{table}

\section{Discussions} 
\label{sec:conclusion}
We introduce a method designed to enable a humanoid robot to track expressive upper body motions while ensuring the maintenance of robust locomotion capabilities in the wild. This method benefits from extensive training on large motion datasets and the use of RSI, equipping the robot with the ability to mimic a wide range of motions responsively and to robustly execute root movement commands that are randomly sampled. Our comprehensive evaluation encompasses both simulated environments and real-world settings. Additionally, the design choices within our framework are rigorously analyzed: quantitatively through simulations and qualitatively through real-world scenarios. We believe our method paves the way for the development of reliable and versatile humanoid robots, capable of performing multiple functions effectively.

\section{Limitations}
In the process of retargeting, the direct mapping of joint angles from the MoCap dataset to the H1 robot, which possesses fewer DoF, leads to a loss of information. Consequently, this can result in the retargeted behavior deviating from the original motion. To mitigate these discrepancies, the application of high-fidelity motion retargeting methods could yield significant improvements. Additionally, given the considerable weight and cost associated with the humanoid robot, the occurrence of falls can sometimes result in broken parts. Therefore, the implementation of a reliable protective and recovery system is imperative to minimize the need for costly repairs.
\clearpage


\bibliographystyle{plainnat}
\bibliography{references}

\clearpage
\def\maketitlesupplementary
   {
   \newpage
       \twocolumn[
        \centering
        \Large
        \textbf{Expressive Whole-Body Control for Humanoid Robots}\\
        \vspace{0.5em} Appendix \\
        \vspace{1.0em}
       ] 
   }
\maketitlesupplementary

\setcounter{section}{0}

\subsection{Motion Dataset Selection}
We curated the training and inference dataset shown in Tab. \ref{tab:dataset} to single person motions on flat terrain to ensure these expressive motions are reasonable to track.
We filter the motions in CMU MoCap by checking if the following keywords are in the description of the motion:
["walk", "navigate", "basketball", "dance", "punch", "fight", "push", "pull", "throw", "catch", "crawl", "wave", "high five", "hug", "drink", "wash", "signal", "balance", "strech", "leg", "bend", "squat", "traffic", "high-five", "low-five"]. And excluding motions with the following keywords: 
["ladder", "suitcase", "uneven", "terrain", "stair", "stairway", "stairwell", "clean", "box", "climb", "backflip", "handstand", "sit", "hang"]. 

\subsection{Additional Training Details}
\noindent \textbf{Rewards}
Expression and root movement goal rewards are specified in Tab. \ref{tab:rewards_detailed}. Regularization reward items are listed in Tab. \ref{tab:regulate_reward}, where $h_\text{feet}$ is feet height, $t_\textit{i}^\text{air}$ indicates the duration each foot remains airborne, $\mathds{1}_\text{new contact}$ represents new foot contact with ground, $\mathbf{F}_\textit{i}^\textit{xy}$ and $F_\textit{i}^\textit{z}$ are for foot contact force in horizontal plane and along the z-axis respectively, with $F_\text{th}$ is the contact force threshold.
$\ddot{\mathbf{q}}$ is joint acceleration, $\mathbf{a_\textit{t}}$ is action at timestep $t$, $\mathds{1}_\text{collision}$ indicates self-collision, $q_\text{max}$ and $q_\text{min}$ are limits for joint positions, $\mathbf{g}_\text{xy}$ is gravity vector projected on horizontal plane.
We specifically add feet related reward items to make sure the feet are comfortably lifted high enough and having a reasonable contact force with the ground when putting down.

\begin{table}[h]
    \centering
\begin{tabular}{@{}llr@{}}
\toprule
Term & Expression & Weight \\
\midrule

\multicolumn{3}{c}{Feet Related} \\ \midrule
Height & max($|\mathbf{h}_\text{feet}|-0.2,0$) & 2.0 \\
Time in Air & $\sum t_\textit{i}^\text{air} * \mathds{1}_\text{new contact}$ & 10.0 \\ 
Drag & $\sum  |\mathbf{v}_i^\text{foot}| * \sim \mathds{1}_\text{new contact} $ & -0.1 \\ 
Contact Force & $ \mathds{1}\left\{ | \textit{F}_\textit{i}^\textit{z} | \geq \textit{F}_\text{th} \right\} * (| \textit{F}_\textit{i}^\textit{z} | - \textit{F}_\text{th}) $ & -3e-3 \\
Stumble & $\mathds{1}\left\{\exists i, \: |\mathbf{F}_\textit{i}^\textit{xy}| > 4 |F_\textit{i}^\textit{z}|\right\}$ & -2.0 \\
\midrule
\multicolumn{3}{c}{Other Items} \\ \midrule

DoF Acceleration & $|\ddot{\mathbf{q}}|^2$ & -3e-7 \\
Action Rate & $|\mathbf{a}_{\textit{t}-1}-\mathbf{a}_\textit{t}|$ & -0.1 \\
Energy & $|\ddot{\mathbf{q}}|^2$ & -1e-3 \\
Collision & $\mathds{1}_\text{collision}$ & -0.1 \\ 
DoF Limit Violation & $\mathds{1}_{q_\textit{i}>q_\text{max}||q_\textit{i}<q_\text{min}}$ & -10.0 \\
DoF Deviation & $|\mathbf{q}_\text{default}^\text{low}-\mathbf{q}^\text{low}|^2$ & -10.0 \\ 
Vertical Linear Velocity & $v_\textit{z}^2$ & -1.0 \\
Horizontal Angular Velocity & $|\boldsymbol{\omega}_\textit{xy}|^2$ & -0.4 \\ 
Projected Gravity & $|\mathbf{g}_\textit{xy}|^2$ & -2.0 \\
\bottomrule
    \end{tabular}
    \caption{Regularization Rewards Specification}
    \label{tab:regulate_reward}
\end{table}
\noindent \textbf{Training Parameters}
We use PPO with hyperparameters listed in Tab. \ref{tbl:ppo_hparams} to train the policy. AMP baseline parameters used in Section \ref{sec:results} are provided in Tab. \ref{tbl:amp_hparams}.
\begin{table}[h]
        \centering
        \raisebox{-0.0\height}{
        \small
        \begin{tabular}{lr}
        \toprule
        Hyperparameter & Value \\ [0.5ex]
        \midrule
         Discount Factor & $0.99$ \\ [0.5ex]
         GAE Parameter & $0.95$ \\ [0.5ex]
          Timesteps per Rollout & $21$ \\ [0.5ex]
          Epochs per Rollout & $5$ \\ [0.5ex]
          Minibatches per Epoch & $4$ \\ [0.5ex]
          Entropy Bonus ($\alpha_2$) & $0.01$ \\ [0.5ex]
          Value Loss Coefficient ($\alpha_1$) & $1.0$ \\ [0.5ex]
          Clip Range & $0.2$ \\ [0.5ex]
          Reward Normalization & yes \\ [0.5ex]
          Learning Rate & $1\mathrm{e}{-3}$ \\ [0.5ex]
          \# Environments & $4096$ \\ [0.5ex]
          Optimizer & Adam \\ [0.5ex]
         \bottomrule
        \end{tabular}
        }
        \vspace{0.45cm}
        \captionof{table}
          {%
            PPO hyperparameters. 
        \label{tbl:ppo_hparams}
          }
\end{table}

\begin{table}[h]
        \centering
        \raisebox{-0.0\height}{
        \small
        \begin{tabular}{lr}
        \toprule
        Hyperparameter & Value \\ [0.5ex]
        \midrule
         Discriminator Hidden Layer Dim & $[1024,512]$ \\ [0.5ex]
         Replay Buffer Size & $1000000$ \\ [0.5ex]
         Demo Buffer Size & $200000$ \\ [0.5ex]
         Demo Fetch Batch Size & $512$ \\ [0.5ex]
         Learning Batch Size & $4096$ \\ [0.5ex]
         Learning Rate & $1e-4$ \\ [0.5ex]
         Reward Coefficient & $4.0$ \\ [0.5ex]
         Gradient Penalty Coefficient & $1.0$ \\
         \bottomrule
        \end{tabular}
        }
        \vspace{0.45cm}
        \captionof{table}
          {%
            AMP hyperparameters. 
        \label{tbl:amp_hparams}
          }
\end{table}
\noindent \textbf{Text2motion Diffusion Model}
We utilize the pre-trained transformer based MDM model to generate human motions from text prompts. In each generating process, 10 repetitions are requested and the most reasonable motion is manually selected for retargetting.

\subsection{Additional Real World Results Visualization}
We provide detailed visualization for some motions evaluated in the real world. Fig. \ref{fig:motion_sup} presents 8 motions from CMU MoCap and 2 motions from text2motion diffusion model. The diffusion model target motions are first generated through MDM \cite{tevet2022human} on SMPL skeleton, then we retarget this motion  to H1 morphology offline. The top images in (k) and (l) are visualizations of target motions rendered with SMPL mesh in Blender. 

\twocolumn[{%
\renewcommand\twocolumn[1][]{#1}%
\begin{center}
    \centering
    \captionsetup{type=figure}
    \includegraphics[width=1.0\textwidth]{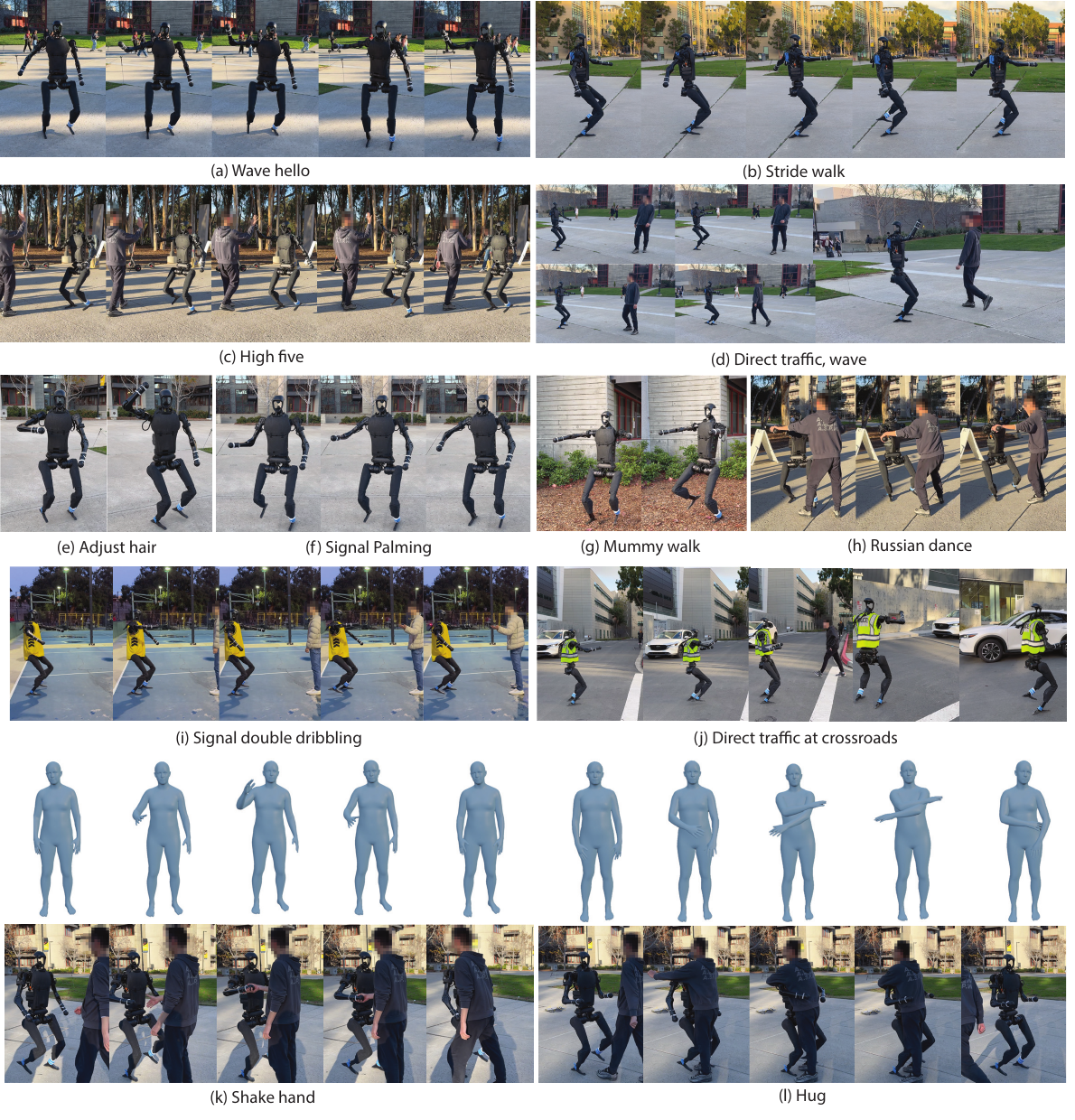}
    \caption{Expressive motion evaluation in the real world. Target motions of (a)-(j) are from CMU MoCap. Target motions of (k) and (l) are prompted using MDM \cite{tevet2022human}. The prompts respectively are "moving arm out to shake hands" and "a person crosses their arms and then puts them back to their side".}
    \vspace{-2pt}
    \label{fig:motion_sup}
\end{center}
}]

\end{document}